\documentclass{article}
\usepackage{arxiv,times}
\usepackage{natbib}
\usepackage[utf8]{inputenc}
\usepackage[T1]{fontenc}
\usepackage{hyperref}
\hypersetup{hidelinks}
\usepackage{url}
\usepackage{graphicx}
\usepackage{booktabs}
\usepackage{amsfonts}
\usepackage{nicefrac}
\usepackage{microtype}
\usepackage{xcolor}
\usepackage{amssymb}
\usepackage{multirow}
\usepackage{enumitem}
\usepackage{amsmath}
\usepackage{algorithm}
\usepackage{algpseudocode}
\usepackage{svg}
\title{ITSY: Causal Discovery From Irregular Time-Series Data}
\author{
Wenbo Xu$^{1}$, Yue He$^{1}$, Yunhai Wang$^{1}$, Yueguo Chen$^{1}$, Kun Kuang$^{2}$\\
\\
$^{1}$Renmin University of China, China \\
$^{2}$Zhejiang University, China \\
}
\begin{document}
\maketitle
\begingroup
\renewcommand{\thefootnote}{}
\footnotetext{Contact: Wenbo Xu (\url{feifeixwb@gmail.com}) and Yue He (\url{hy865865@gmail.com}).}
\addtocounter{footnote}{-1}
\endgroup

\maketitle

\begin{abstract}
Structural causal models for time series recover contemporaneous and lagged effects, but most methods require complete observation windows and become misspecified when samples are missing.
We introduce \textbf{ITSY}, the first continuous-optimization method for causal discovery from irregular time series under a linear model. 
ITSY reformulates the structural equation so that prediction uses the nearest available history rather than the possibly missing current slice, and jointly imputes missing values while learning both graphs.
A weighted reconstruction objective corrects the noise transformation induced by this reformulation. 
Across synthetic regimes varying missingness, scale, graph density, and noise, and on a real world benchmark, ITSY consistently improves graph recovery over representative SCM-based baselines, demonstrating the effectiveness of the proposed method.
The results establish a focused solution for irregular linear first-order dynamics and clarify the assumptions required for nonlinear or higher-order extensions.
\end{abstract}

\section{Introduction}
Causal discovery represents dependencies with graphs and supports scientific analysis in biology, environmental science, agriculture, and economics \citep{pcmci2020bio,pcmci2021geo,pcmci2025agriculture,granger2021economic}. 
Beyond them, temporal discovery is particularly difficult because effects may occur both within a time slice and across lags.

To address such issues, different kinds of algorithms have been introduced. 
Granger causality \citep{granger1969investigating} is determined by the improvement of prediction, while Ordinary Differential Equation (ODE) \citep{ODE2018neural} model dependency relationships based on the changing rate in continuous dynamics.
Although Granger causality and ODE-based methods have been widely applied, both methods ignore contemporaneous effects, only focus on past data.
Clearly, this assumption is limited compared with real scenarios.
For example, the current price of a stock depends on both other contemporaneous price and the past ones.

Structural Causal Models (SCM) proposed by \citet{pearl2009causality} explicitly defines structural equations as generating processes of different variables and free causal analysis from relying on past data alone. 
Under temporal settings, SCM treats variables at each time step as a separate node in the causal graph, including contemporaneous effect.
Recent years, continuous SCM estimators have made graph learning scalable \citep{NOTEARS2018dags,zheng2020NOTEARSMLP,gong2024rhino,sun2023nts,sanchez2023diffusion}, but they typically assume that every required time slice is observed.
This assumption fails under sensor dropouts, asynchronous acquisition, and transmission loss.
A separate statistical or neural imputer \citep{cao2018brits} can fill the gaps, but its prediction objective need not preserve the causal generating mechanism and may bias the recovered graph \citep{cuts2023cheng}.

To mend this gap, we study recovery of contemporaneous and lagged graphs from irregular observations. 
Our method, \textbf{I}rregular \textbf{T}ime-\textbf{S}eries Causal Discover\textbf{Y} (\textbf{ITSY}), first shows why standard SVAR reconstruction fails when the current slice is missing.
It then regroups the contemporaneous terms and predicts from the nearest available history. 
Imputation and graph learning are optimized jointly with reconstruction, sparsity, and acyclicity objectives. 
The model is deliberately scoped to linear first-order SVARs: this setting makes the contemporaneous/lagged decomposition explicit and permits an exact weighted-loss correction.
Nonlinear mechanisms require a structured invertible contemporaneous map, whereas higher-order dynamics require an augmented lag state;
we state these boundaries instead of attributing untested generality to ITSY.

The central difficulty is not merely to complete the data matrix, but to preserve the distinction between effects within the current slice and effects arriving from the past. 
A generic imputer \citep{cao2018brits} can produce plausible values while changing the residual structure used to estimate these two graphs.
ITSY instead implicitly derives each missing value from the same structural transition that defines graph learning. 
This mechanism-level coupling lets observed entries supervise both reconstruction and causal parameters, while the mask prevents recursively imputed values from being treated as direct observations.
In summary, our main contributions are as follows:
\begin{itemize}[leftmargin=9pt]
\item We formulate irregular temporal causal discovery for SCMs containing both contemporaneous and
lagged effects, and characterize the failure of complete-window reconstruction under missingness.
\item We develop a continuous objective that alternates mechanism-aware imputation with graph learning
and uses weighted residuals to counteract the noise transformation caused by SVAR reformulation.
\item We evaluate ITSY across diverse synthetic settings and a semi-synthetic fMRI benchmark, showing
consistent improvements under the stated linear, first-order assumptions.
\end{itemize}

\section{Related Work}
\subsection{Granger and ODE based methods}
Recent years have witnessed numerous works integrating Granger causality with neural networks, including CNNs, LSTMs, and VAEs
\citep{TCDF2019causal,NGC2021granger,esru2020,cVAE2023causal}.
\citet{cuts2023cheng,cuts+2024cheng} develop a two-stage iterative approach to jointly impute missing data points and infer Granger causality.
However, such predictive causality focuses only on past observations and cannot model contemporaneous effects.

Neural ODEs, proposed by \citet{ODE2018neural}, model temporal dynamics as a continuous system by treating a neural network as the differential function.
Based on this framework, \citet{ODE2020pair} provide a method for identifying pairwise causality, while \citet{ODE2022causal} model the dependence graph of a dynamic system from multivariate time series.
\citet{SDE2024causal} extend these approaches to more general settings by using stochastic differential equations.
\citet{cheng2025dycast} provide an initial exploration of temporal causal discovery with dynamically evolving graph structures.
CADYT \citep{tagliapietra2026causal} performs explicit graph search using ODE with a theoretically motivated score.
Although ODE-based methods above can handle irregularity without imputation, their theoretical foundations generally ignore contemporaneous interactions.

\subsection{SCM-based methods}
SCM-based methods incorporate contemporaneous effects into their structural equations.
Constraint-based methods model causal graphs through conditional independence among time series \citep{pcmci2019runge,pcmci+2020runge,RPCMCI2020saggioro,PCGCE2022discovery,OCSE2015causal}.
Function-based methods identify causal structures through predefined structural equations with specific noise forms~\citep{varlingam2010causal, timino2013causal,NBCB2021mixed,NBCB2024causal,NCDH2022nonlinear}.
Score-based methods identify graph structures through score functions and NOTEARS transforms it into a differentiable continuous optimization problem \citep{NOTEARS2018dags,zheng2020NOTEARSMLP}.
Along this line, DYNOTEARS achieves linear temporal causal discovery
\citep{pamfil2020dynotears}, while NTS-NOTEARS \citep{sun2023nts} extends the framework to nonlinear time series.
RHINO \citep{gong2024rhino}, on the other hand, models temporal causal relationships through variational inference.
Despite their flexibility, these methods assume fully observed data and may therefore lead to model misspecification in real-world scenarios with irregular observations.
We do not consider standalone neural or statistical imputation methods because they ignore the underlying causal mechanisms.
Recent years, \citet{gao2022missdag} explore missingness in non-temporal settings.
ReTimeCausal~\citep{li2026causaldiscoveryirregularlytime} jointly performs imputation and causal discovery through an EM-style procedure, but is restricted to lagged summary graphs and limits to model contemporaneous effects.

To this end, we propose ITSY to improve the applicability of SCM-based methods in real-world scenarios by modeling both lagged and contemporaneous causal effects in irregular time series.
By learning the generating mechanism, ITSY imputes missing values according to the estimated data-generating process.

\section{Preliminaries}
\subsection{Dynamic Causal Modeling}

\begin{figure*}[t]
    \centering
    \includegraphics[width=1.0\textwidth]{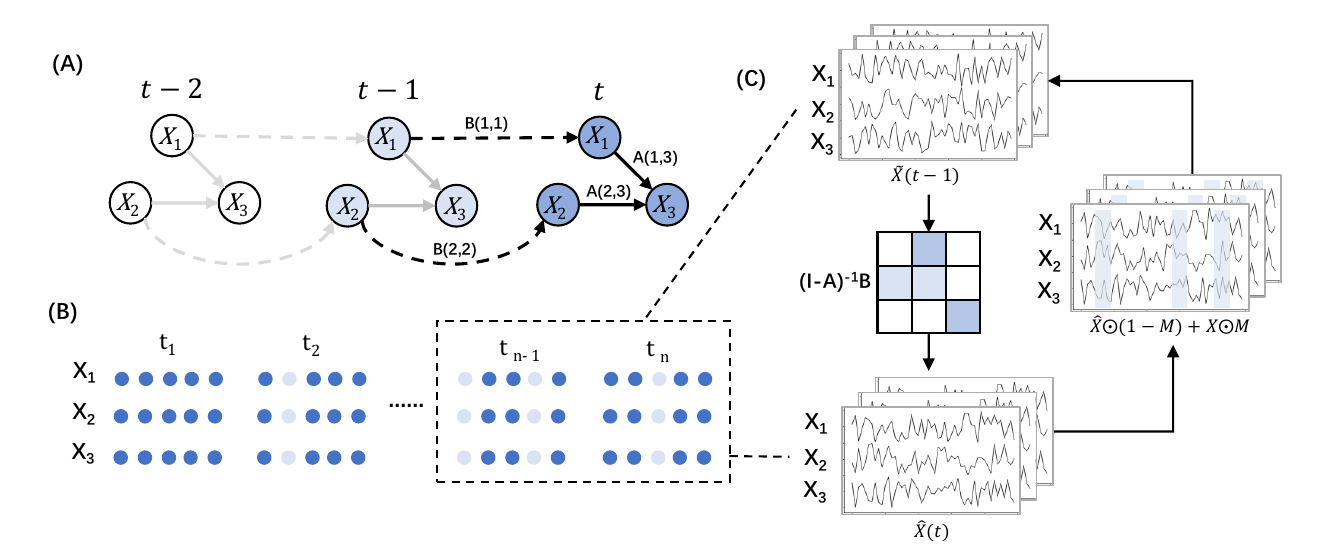}
    \caption{(A) SVAR model. Solid lines denote contemporaneous weights, dashed lines denote past weights, and gray lines show weights do not affect $\mathbf{X}(t)$. $(m,n)$ indicates indices of weight matrices $\mathbf{A}$ or $\mathbf{B}$, representing $\mathbf{X}_m$ has influence on $\mathbf{X}_n$ temporally.
    (B) Irregular Patterns, where gray indicates missing points.
    (C) Model Pipeline. Prediction and imputation are conducted iteratively.
    }
    \label{fig1}
\end{figure*}
\paragraph{SVAR}  
We model the time series with both weights using the following standard structural vector autoregression (SVAR) model, which is proposed by \citet{svar1997impulse}.
\begin{equation}\label{eq1}\small
\mathbf{X}(t) = \mathbf{AX}(t)+\sum_{\tau=1}^{k} \mathbf{B}_\tau \mathbf{X}(t-\tau) + \mathbf{e}(t)
\end{equation}
where $k$ is the order, $\mathbf{A}$ and $\mathbf{B}_\tau$ are contemporaneous and lagged weight matrices, and $\mathbf{e}(t)$ contains errors independent across variables and time.
The tensor $\mathbf{X}\in\mathbb{R}^{T\times n\times d}$ contains $d$ variables, $T$ time steps, and $n$ samples.
ITSY uses $k=1$ (Fig.~\ref{fig1}(A)); $k>1$ requires augmented-state completion and is left for future work.

\paragraph{Causal Modeling} 
An SCM consists of a directed acyclic graph (DAG) $G=(V,E)$ and structural equations, with $P(\mathbf{X})=\prod_i P(x_i\mid\mathbf{X}_{pa(i)})$. Following other temporal SCMs
\citep{varlingam2010causal,pamfil2020dynotears}, $\mathbf{A}$ and $\mathbf{B}_\tau$ encode contemporaneous and lagged effects; only $\mathbf{A}$ requires a DAG constraint because time orders
lagged edges. 
We assume causal sufficiency:

\textit{Assumption 1 (Causal Sufficiency)} \citep{pearl2009causality}: $\mathbf{X}$ contains every
common cause of the observed variables, so the causal structure is represented by a DAG without
latent confounders.

\textit{Assumption 2 (Causal Stationarity)}\citep{gong2024causal}: the structural functions and
parent sets are time homogeneous. Thus, at every $t$, $x_i(t)=f_i(\mathbf{X}_{pa(i)}(t))+e_i(t)$ with the same $f_i$.

\textit{Assumption 3 (Ignorable Missingness)}\citep{rubin1976inference}:
 we assume that missingness is ignorable, thus our model considers the Missing Completely at Random (MCAR) setting, where the probability that an entry is missing is independent of both observed and unobserved values.
 
As said by \citet{gong2024causal}, we use \textbf{Window Graph} to model causality.
Granger methods usually return a \textbf{Summary Graph}, $G\in\mathbb{R}^{d\times d}$ in which $x_i(t-)\rightarrow x_j(t)$ denotes predictive information from the history of $x_i$.
Similarly, ODE-based methods also use \textbf{Summary Graph} but to parameterize changing rates.
SCM uses a \textbf{Window Graph} of $G\in\mathbb{R}^{\tau \times d\times d}$: an edge $x_i(t-\tau)\rightarrow x_j(t)$ identifies a structural mechanism at lag $\tau\in\ N$ and $\tau=0$ means contemporaneous effect.
ITSY follows SCM and the \textbf{Window Graph} paradigm, and thus constructs multiple causal graphs capturing both contemporaneous and time-lagged causal relationships.

\subsection{Irregularity in time series}
Based on empirical observations, we addresses two types of irregularity in time series data $\mathbf{X}$ . 
\textbf{Sample-wise Missing} indicates each individual observation $\mathbf{X}^i(t)$ at time $t$ is independently missing with probability $p$ and yields $\lfloor p\,n\rfloor \times d$ missing entries in per observations $\mathbf{X}(t)$, see Fig.\ref{fig1}(B).
\textbf{Step-wise Missing} arises when the observation vector $\mathbf{X}(t_i)$ is entirely absent at time step $t_i$ with probability $p$, resulting in $\lfloor p\,T\rfloor \times n \times d$ missing samples overall, see Tab.\ref{tab1}.
Both $\mathbf{X}^i(t)$ and $\mathbf{X}(t_i)$ follow the $\sim Bernoulli(1 - p)$, so that $\mathbf{X}$ are retained with probability $(1-p)$.

There is a special case called Subsamping, where data is sampled at a fixed interval, i.e., $\{\textbf{X}(1), \textbf{X}(1+a),\ldots ,\textbf{X}(1+a(T-1))\}$.
\citet{gong2015subsample} and \citet{liu2023subsample} theoretically highlight the challenges of recovering causal relations under this setting.
By contrast, we employ a more general setting of $p<0.5$ in the experiments, which ensures sufficient samples without interval.

\section{Motivation}
\label{chap4}

\begin{table*}[t]
    \centering
    \includegraphics[width=1.0\linewidth]{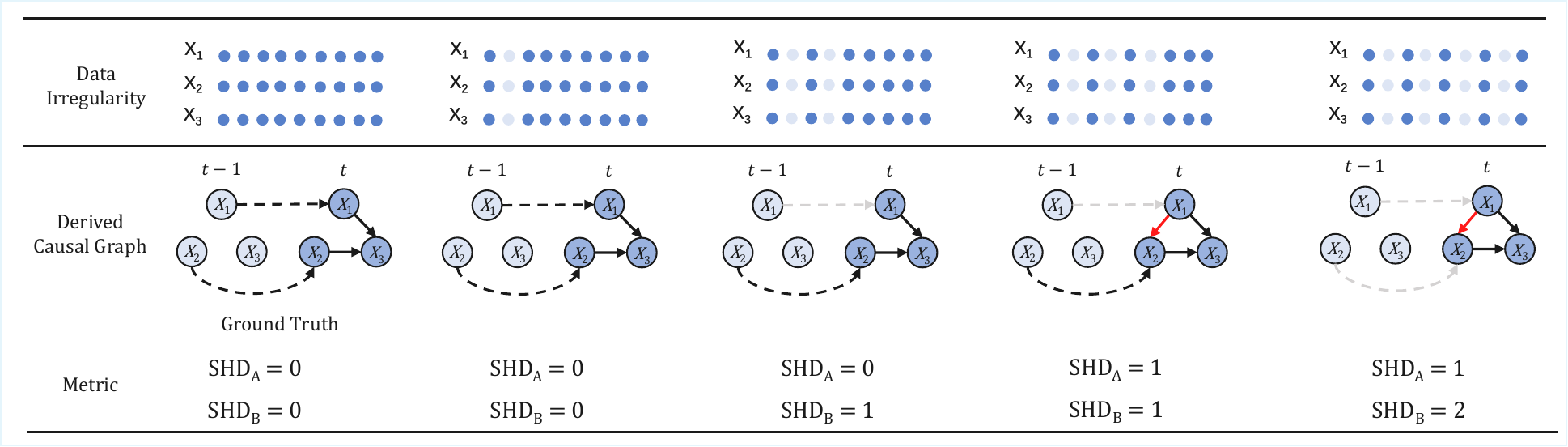}
    \caption{Toy example of step-wise missingness in temporal causal discovery. Red and gray edges denote spurious and missed edges, respectively; lower SHD is better.}
    \label{tab1}
\end{table*}

Temporal SCM methods generally require complete and regularly observations, and is violated when irregularity happens.
Direct deletion changes effective time intervals, while statistical imputation can alter causal signals.
We therefore preserve the temporal grid with zero-padding and use DYNOTEARS \citep{pamfil2020dynotears} as a toy example.

With complete observations, DYNOTEARS reconstructs both causal graphs with zero SHD.
When $\mathbf{X}(1)$ is padded, $\hat{\mathbf{X}}(1)=\hat{\mathbf{B}}\mathbf{X}(0)$ omits the contemporaneous contribution, while $\hat{\mathbf{X}}(2)=\hat{\mathbf{A}}\mathbf{X}(2)$ omits the lagged contribution.
Minimizing reconstruction loss under these incomplete equations provides misleading supervision for both $\hat{\mathbf{A}}$ and $\hat{\mathbf{B}}$.
The lagged matrix is optimized toward the padded target and consequently tends to shrink, whereas the contemporaneous matrix compensates for the missing lagged contribution.
At this stage, the remaining complete temporal windows still provide sufficient supervision, allowing both graphs to be recovered with zero SHD.
When $t=1$ and $t=3$ are missing, the same distortion propagates to the subsequent predictions at $t=2$ and $t=4$.
Nearly half of the relevant reconstruction terms are then misspecified, causing $\hat{\mathbf{B}}$ to lose the edge $\hat{\mathbf{B}}(1,1)$.
After the observation at $t=5$ is removed, fewer complete transitions remain and $\hat{\mathbf{A}}$ begins to overfit the residual contemporaneous signal.
This compensation introduces the spurious edge $\hat{\mathbf{A}}(1,2)$ even though the corresponding relationship is absent from the generating graph.
When all odd-indexed observations are missing, the sequence becomes effectively subsampled and each estimate is supported primarily by either the contemporaneous or lagged mechanism.
Consequently, $\hat{\mathbf{B}}$ becomes incomplete, $\hat{\mathbf{A}}$ overfits the remaining observations, and the SHDs of both graphs deteriorate.

Tab.~\ref{tab1} shows that irregularity can systematically bias the objectives used for temporal causal discovery.
This issue extends beyond DYNOTEARS: function-based methods require aligned observations, while constraint-based methods require jointly observed variables for conditional-independence tests.
ITSY instead couples mechanism-aware completion with graph learning, allowing reconstructed states and structural graphs to constrain one another.
\clearpage
\section{Proposed Method: ITSY}
\subsection{Causal Process}
With complete observations, continuous SCM methods learn structural equations and graph parameters end-to-end \citep{zheng2020NOTEARSMLP,pamfil2020dynotears,varlingam2010causal}.
For the first-order SVAR in Eq.~\ref{eq1}, the usual one-step reconstruction is:
\begin{equation} \small \label{eq2}
    \hat{\mathbf{X}}(t) = \hat{\mathbf{A}}\mathbf{X}(t) + \hat{\mathbf{B}}\mathbf{X}(t-1),
\end{equation}
This reconstruction underlies the least-squares formulation of DYNOTEARS \citep{pamfil2020dynotears}.
Eq.\ref{eq2}, however, uses the current observation on both sides and is no longer applicable in the presence of missing data.
Rearranging the structural equation gives:
\begin{equation} \small \label{eq3}
    (\mathbf{I}-\mathbf{A})\mathbf{X}(t) = \mathbf{B}\mathbf{X}(t-1) + \mathbf{e}(t),
\end{equation}
Under the acyclicity constraint, $(\mathbf{I}-\mathbf{A})$ is invertible, as proof shown in Appendix~\ref{A_A_proof}.
Let $\mathbf{P}_{\theta}=(\mathbf{I}-\mathbf{A})^{-1}\mathbf{B}$ denote the induced transition operator.
The learned mechanism then predicts:
\begin{align} \small \label{eq4}
 (\mathbf{I}-\hat{\mathbf{A}})\hat{\mathbf{X}}(t) = \hat{\mathbf{B}}\mathbf{X}(t-1)
 \quad\Rightarrow\quad &\hat{\mathbf{X}}(t)=\mathbf{P}_{\hat{\theta}}\mathbf{X}(t-1).
\end{align}
Unlike previous equation, Eq.~\ref{eq4} predicts a current state without taking that state as input.
For two missing slices, $\hat{\mathbf{X}}(t)=\hat{\mathbf{P}}\hat{\mathbf{X}}(t-1)=\hat{\mathbf{P}}^{2}\mathbf{X}(t-2)$.
If the nearest preceding observation is $i$ steps away, the general form is:
\begin{equation}\label{eq6}
    \hat{\mathbf{X}}(t)=\mathbf{P}_{\hat{\theta}}^{i}\mathbf{X}(t-i).
\end{equation}
Each application of $\mathbf{P}_{\hat{\theta}}$ represents one SVAR transition, so $\mathbf{P}_{\hat{\theta}}^{i}$ propagates the last available state across exactly $i$ missing steps.
The same graph parameters therefore govern both observed and unobserved intervals without introducing gap-specific transition operators.
Thus, irregular sampling is handled by repeatedly applying the same causal transition over missing steps, without redefining the lag structure.
This mechanism-level parameter sharing is the basis for the joint formulation below.

\subsection{Joint Optimization Problem}
We introduce \textbf{I}rregular \textbf{T}ime-\textbf{S}eries Causal Discover\textbf{Y} (\textbf{ITSY}) to handle missingness.
ITSY treats the missing values as implicit functions of the learned graph parameters rather than as outputs of a separate preprocessing stage.
To distinguish the irregular input $\mathbf{X}$ from its completed version, we use $\tilde{\mathbf{X}}_{\theta}$ for the full model-completed series, which retains observed entries and replaces only missing entries with the causal predictions $\hat{\mathbf{X}}_{\theta}$.

Let $\mathbf{M}(t)\in\{0,1\}^{n\times d}$ denote the observation masks, which exclude unobserved entries from supervision. 
A forward application of the causal operator defines:
\begin{align}\small\label{eq7}
\hat{\mathbf{X}}_{\theta}(t)&=\mathbf{P}_{\theta}\tilde{\mathbf{X}}_{\theta}(t-1), \nonumber\\
\tilde{\mathbf{X}}_{\theta}(t)&=\mathbf{M}(t)\odot\mathbf{X}(t)
+(1-\mathbf{M}(t))\odot\hat{\mathbf{X}}_{\theta}(t).
\end{align}
Eq.\ref{eq7} clamps every observation to its measured value and fills only unobserved coordinates.
Because $\mathbf{P}_{\theta}=(\mathbf{I}-\mathbf{A})^{-1}\mathbf{B}$, changing either the contemporaneous graph $\mathbf{A}$ or the lagged graph $\mathbf{B}$ alters the downstream completed values and hence affects the structural evidence used to update both graphs.
This feedback motivates coupling data completion and graph learning within a reduced joint program rather than a sequential pipeline.
Since Eq.~\ref{eq7} recursively determines $\tilde{\mathbf{X}}_{\theta}$ from $(\mathbf{X}$, $\mathbf{M})$, and $\theta=(\mathbf{A},\mathbf{B})$, the completed series is not an independent optimization variable.

ITSY therefore minimizes the following observed-entry prediction loss over $\mathbf{A}$ and $\mathbf{B}$:
\begin{align}\small\label{eq8}
&\min_{\mathbf{A},\mathbf{B}}\ \mathcal{L}_{rec}(\theta)
\quad\text{s.t.}\quad \mathbf{A}\ \text{is a DAG},\nonumber\\
&\mathcal{L}_{rec}(\theta)=\frac{1}{2n}\sum_{t=1}^{T}
\|\mathbf{M}(t)\odot(\mathbf{X}(t)-\hat{\mathbf{X}}_{\theta}(t))\|_F^2.
\end{align}

The mask $\mathbf{M}$ excludes imputed targets from direct supervision, while their values remain active as contexts for later observed targets.
Consequently, the total gradient contains both a direct graph-fitting term and an indirect completion-feedback term:
\begin{equation}\small\label{eq_joint_gradient}
\frac{d\mathcal{L}_{rec}}{d\theta}
=\frac{\partial\mathcal{L}_{rec}}{\partial\theta}
+\sum_t\frac{\partial\mathcal{L}_{rec}}{\partial\tilde{\mathbf{X}}_{\theta}(t)}
\frac{d\tilde{\mathbf{X}}_{\theta}(t)}{d\theta}.
\end{equation}
In which, the second term formalizes the coupling: graph updates are evaluated through the causal mechanism that produces the imputed states.
This coupling jointly learns completion and structure without introducing an independent imputation model or an EM interpretation.

To promote sparse graph structures, we incorporate element-wise $\ell_1$ penalties into the objective:
\begin{equation}\small
\mathcal{L}_{sparse}(\mathbf{A},\mathbf{B})
=\lambda_A\|\mathbf{A}\|_1+\lambda_B\|\mathbf{B}\|_1.
\end{equation}
Only the contemporaneous graph requires an acyclicity constraint because lagged edges point forward in time.
We use the differentiable trace-exponential characterization of \citet{NOTEARS2018dags}.

\textit{Lemma 1.}
A matrix $\mathbf{G} \in \mathbb{R}^{d \times d}$ represents a DAG if and only if:
\begin{equation} \small
    h(\mathbf{G})=\text{tr}\left(\mathrm{e}^{\mathbf{G} \circ \mathbf{G}}\right) - d = 0.
\end{equation}
This characterization replaces the discrete constraint in Eq.~\ref{eq8} with the equality $h(\mathbf{A})=0$.
Following \citet{NOTEARS2018dags}, we enforce this equality through the augmented Lagrangian:
\begin{equation} \small
    \mathcal{L}^{\rho,\alpha}_{DAG}(\mathbf{A})=\frac{\rho}{2} h(\mathbf{A})^2 + \alpha h(\mathbf{A})
\end{equation}
Appendix \ref{A_A_proof} provides the corresponding derivation.

\begin{algorithm}[t]
\caption{ITSY: joint graph learning and time-series completion}
\label{alg1}
\begin{algorithmic}[1]
\Statex \textbf{Input:} Irregular series $\mathbf{X}$, mask $\mathbf{M}$, $\lambda_A,\lambda_B$
\Statex \textbf{Output:} Contemporaneous graph $\mathbf{A}$, lagged graph $\mathbf{B}$
\State Initialize $\mathbf{A},\mathbf{B}$ and multipliers $\rho,\alpha$
\While{not converged and $\rho<\rho_{\max}$}
    \Repeat
        \State $\mathbf{P}_{\theta}\gets\operatorname{solve}(\mathbf{I}-\mathbf{A},\mathbf{B})$
        \State Set $\tilde{\mathbf{X}}_{\theta}(0)$ from the initial observation
        \For{$t\gets1,\ldots,T$}
            \State $\hat{\mathbf{X}}_{\theta}(t)\gets\mathbf{P}_{\theta}\tilde{\mathbf{X}}_{\theta}(t-1)$
            \State Complete $\tilde{\mathbf{X}}_{\theta}(t)$ by Eq.~\ref{eq7}
        \EndFor
        \State $\mathcal{O}\gets\mathcal{L}'_{rec}+\mathcal{L}_{sparse}+\mathcal{L}^{\rho,\alpha}_{DAG}$
        \State Update $\mathbf{A},\mathbf{B}$ through the differentiable sweep
    \Until{the inner objective converges}
    \State Update $\alpha$ and $\rho$ using the current $h(\mathbf{A})$
\EndWhile
\end{algorithmic}
\end{algorithm}

\subsection{Differentiable Coupled Solver}
Equation~\ref{eq7} is evaluated by a forward sweep inside every optimization step.
The sweep remains in the computation graph, allowing gradients to propagate through the completed values rather than treating them as fixed pseudo-observations.
Numerically, $\mathbf{P}_{\theta}$ is obtained by solving a linear system instead of explicitly forming an inverse.
At inner iteration $q$, the current graphs first construct $\tilde{\mathbf{X}}_{q}$ through Eq.~\ref{eq7} and then follow a descent step on the reduced objective:
\begin{equation}\small\label{eq_solver_update}
\theta_{q+1}=\theta_{q}-\eta_q
\nabla_{\theta}\mathcal{O}\!\left(\theta_q,\tilde{\mathbf{X}}_{q}\right).
\end{equation}
Since $\tilde{\mathbf{X}}_{q}$ depends on $\theta_q$, this gradient differentiates through the imputed values and updates completion and graph parameters in one coupled step.
The outer loop follows the augmented-Lagrangian schedule for acyclicity, updating $\alpha_{s+1}=\alpha_s+\rho_s h(\mathbf{A}_s)$ and increasing $\rho_s$ when violation reduction is insufficient.
The inner loop stops when the objective stabilizes, while the outer loop additionally requires $h(\mathbf{A})$ to approach zero.
Algorithm~\ref{alg1} summarizes the resulting solver.

\subsection{Model Analysis} 
\begin{figure}[t] 
    \centering 
    \includegraphics[width=\textwidth]{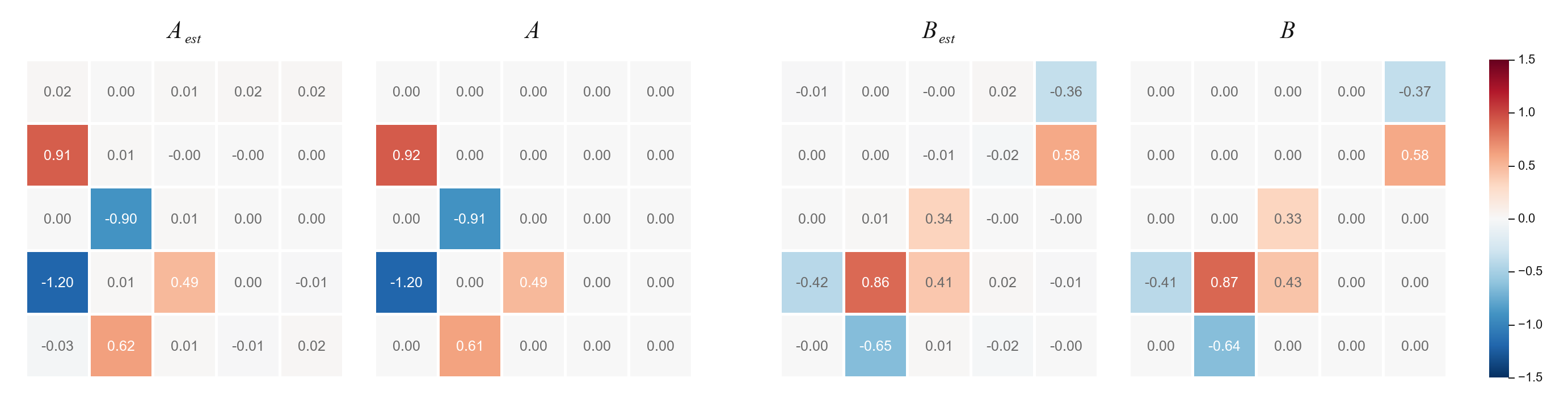}
    \caption{Illustration with Gaussian noise, ITSY recovers weights close to the ground truth.
    } 
    \label{figprob} 
\end{figure}
\citet{ng2024sober} and \citet{he2021daring} provide empirical evidence that continuous structure-learning methods are sensitive to unequal noise variances.
\citet{ng2024sober} further show that the nonconvexity of continuous optimization objectives can produce suboptimal local solutions.
We examine this limitation from the perspective of reconstruction loss and identify how the structural transformation alters the effective noise distribution.

\textit{Lemma 2 (Gauss--Markov theorem).}\label{lemma2}
Under zero-mean, homoscedastic, and uncorrelated errors, ordinary least squares (OLS) is the best linear unbiased estimator (BLUE) \citep{casella2002statistical}.
Appendix \ref{A_A_proof} provides the derivation used in our analysis.

Although Eq.~\ref{eq1} assumes independent errors $\mathbf{e}(t)\sim(0,\sigma^2)$ and therefore satisfies the conditions of Lemma 2, the prediction form in Eq.~\ref{eq4} transforms the innovation into $\mathbf{u}(t)=(\mathbf{I}-\mathbf{A})^{-1}\mathbf{e}(t)$.
The inverse exists because a DAG adjacency matrix is nilpotent, hence $\mathbf{I}-\mathbf{A}$ is nonsingular.
The transformed residual remains zero-mean, but its covariance $\operatorname{Cov}[\mathbf{u}(t)]=\sigma^2(\mathbf{I}-\mathbf{A})^{-1}(\mathbf{I}-\mathbf{A})^{-\top}$ is generally neither diagonal nor homoscedastic.
Direct OLS on the prediction residual therefore violates the conditions that make it BLUE and can favor a graph whose inverse reshapes the noise into an easier but wrong regression target.

We remove this geometry-dependent preference with weighted least squares (WLS).
Premultiplying the prediction residual by $\mathbf{I}-\mathbf{A}$ whitens it in structural coordinates because $(\mathbf{I}-\mathbf{A})\mathbf{u}(t)=\mathbf{e}(t)$.
Applied to the completed series defined by Eq.~\ref{eq7}, the corrected objective is:
\begin{equation} \small \label{eq12}
\mathcal{L'}_{rec}(\theta)=\frac{1}{2n}\sum_{t=1}^{T}
\left\|\mathbf{M}(t)\odot(\mathbf{I}-\mathbf{A})
\big(\tilde{\mathbf{X}}_{\theta}(t)-\hat{\mathbf{X}}_{\theta}(t)\big)\right\|_F^2.
\end{equation}
The correction is exact for the linear model because $\hat{\mathbf{X}}_{\theta}(t)=\mathbf{P}_{\theta}\tilde{\mathbf{X}}_{\theta}(t-1)$ makes the quantity inside the mask equal to $(\mathbf{I}-\mathbf{A})\tilde{\mathbf{X}}_{\theta}(t)-\mathbf{B}\tilde{\mathbf{X}}_{\theta}(t-1)$, i.e., the original structural residual.
The mask then selects equations with observed targets rather than asking recursively generated values to validate themselves.
Appendix~\ref{A_B_coverge} gives the derivation and implementation details.
With $\tilde{\mathbf{X}}_{\theta}$ determined by Eq.~\ref{eq7}, combining structural weighting, sparsity, and acyclicity yields the final reduced problem:
\begin{equation} \small\label{eq_final_objective}
\min_{\mathbf{A},\mathbf{B}}\ \mathcal{L}(\theta)
=\mathcal{L'}_{rec}(\theta)+\mathcal{L}_{sparse}(\mathbf{A},\mathbf{B})
+\mathcal{L}^{\rho,\alpha}_{DAG}(\mathbf{A}).
\end{equation}
Eq.\ref{eq_final_objective} couples data completion with structural weighting: missing entries follow the learned mechanism, while WLS evaluates contemporaneous and lagged effects together.

For a run of $i$ missing slices, Eq.~\ref{eq6} contributes through $\mathbf{P}_{\theta}^{i}$, whose differential is $d(\mathbf{P}_{\theta}^{i})=(d\mathbf{P}_{\theta})\mathbf{P}_{\theta}^{i-1}+\cdots+\mathbf{P}_{\theta}^{i-1}(d\mathbf{P}_{\theta})$.
Thus, a later observed error supervises all intervening causal paths, although long gaps may amplify vanishing or exploding powers of $\mathbf{P}_{\theta}$; observed entries in Eq.~\ref{eq7} reset this recursion.
One coupled sweep costs $O(d^3+Tnd^2)$ time and $O(Tnd)$ memory: a linear solve defines $\mathbf{P}_{\theta}$, followed by differentiable propagation without a separate imputation network or per-entry parameters.
Further details are provided in the Appendix \ref{A_B_coverge}.

The analysis is specific to linear first-order SVARs; nonlinear mechanisms require state-dependent inverse maps and additional derivative terms, while higher-order dynamics require augmented states and new identification and stability arguments.

\section{Experiments} 
\subsection{Experimental Settings}

\paragraph{Baselines and metrics.}
We compare ITSY with three SCM estimators that recover contemporaneous and lagged edges under compatible graph semantics.
\citet{pcmci+2020runge} introduce\textbf{ PCMCI+}, which identifies temporal relations through conditional-independence tests across variables and time.
\citet{pamfil2020dynotears} formulate \textbf{DYNOTEARS} as a continuous optimizer for linear SVARs, while \citet{sun2023nts} and \citet{gong2024rhino} provide \textbf{NTS-NOTEARS} and \textbf{Rhino} to represent nonlinear temporal mechanisms.
\textbf{CUTS} and \textbf{CUTS+}~\citep{cuts2023cheng,cuts+2024cheng} also process irregular samples, but their outputs are Granger graphs rather than the two SCM graphs evaluated here.
We omit function-based estimators because their stronger noise assumptions and different scalability regimes make a controlled comparison hard.
Appendix \ref{A_C_baseline} reports all detailed implementations, thresholds, and hyperparameters.
We evaluate graph recovery using \textbf{F1}, where higher is better, and \textbf{Structural Hamming Distance (SHD)}, where lower is better.
F1 summarizes the balance between edge precision and recall, whereas SHD counts missing, reversed, and additional edges.
Reporting both metrics prevents a sparse but incomplete graph from appearing preferable solely because it requires fewer total edits.
All missing masks are sampled independently of the simulated values, corresponding to an MCAR-like observation process rather than MAR or MNAR missingness.
 
\subsection{Synthetic Data}
\begin{figure}[t]
\centering
\includegraphics[width=\textwidth]{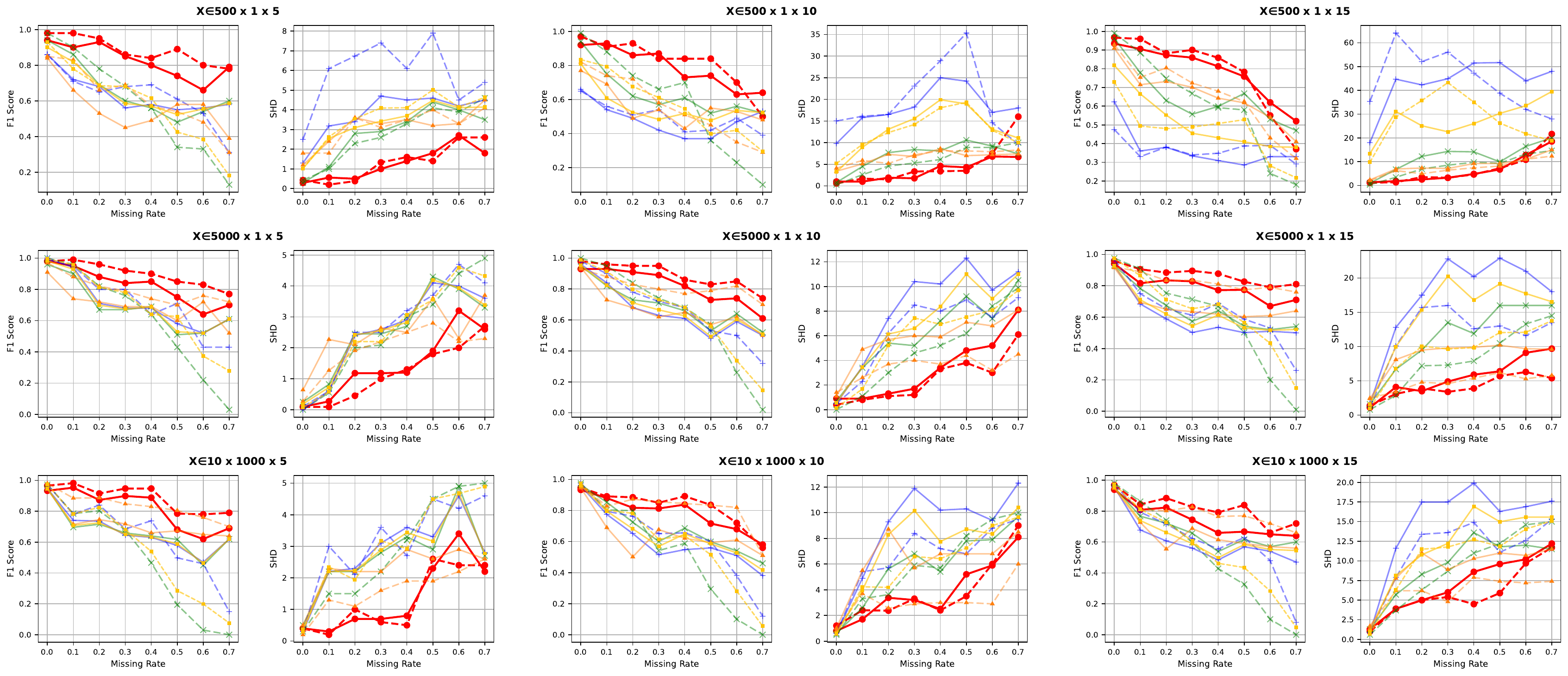}

\includegraphics[width=0.60\textwidth]{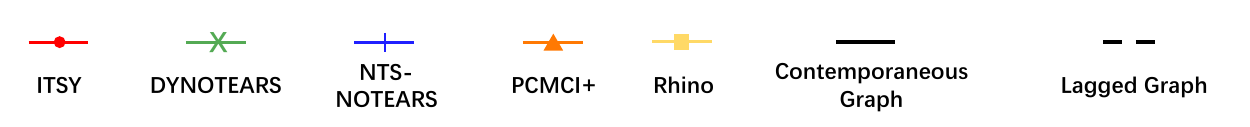}

\caption{Graph recovery across scales. Rows vary $d\in\{5,10,15\}$, while columns vary sequence length and dimension. Colors and types denote methods and lag or not.}
\label{fig2}
\end{figure}

We generate data from a first-order SVAR with $\mathbf{e}(t)\sim\mathcal{N}(0,1)$.
The Erd\H{o}s--R\'enyi graphs $(\mathbf{A},\mathbf{B})$ have average degree $(1,1)$ and nonzero weights in $(-0.95,-0.5)\cup(0.5,0.95)$.
Missing rates range from 0 to 70\%, with step-wise missingness for $n=1$ and sample-wise missingness for replicated series.
Fig.\ref{fig2} varies the sequence length, dimension, and number of aligned series while reporting F1 and SHD throughout the missingness range.
At zero missingness, NTS-NOTEARS and PCMCI+ become competitive when sufficient observations are available, such as $T=5000$ or $n=1000$.
These complete-data results serve as a consistency check and do not imply that ITSY universally dominates estimators designed for fully observed sequences.
As missingness increases, the baselines deteriorate more rapidly, whereas ITSY retains higher F1 and lower SHD across the tested scales.
The advantage appears in both contemporaneous and lagged recovery, indicating that recursively completing temporal windows also protects estimation of the current-time structure.
Appendix \ref{A_D_Exp} varies the noise distribution and graph density for $\mathbf{X}\in\mathbb{R}^{5000\times1\times10}$ and additionally considers $d\in\{5,15\}$.
ITSY remains competitive for sparse $(1,0)/(0,1)$ and denser $(1,2)/(2,1)$ graphs, supporting robustness within the linear first-order model family.
Appendix \ref{A_B_coverge} further reports convergence behavior and runtime.
All methods receive identical generating sequences and missing masks, and every estimate is evaluated against the unchanged ground-truth graphs.
The widening performance gap therefore isolates tolerance to incomplete reconstruction windows rather than differences in data generation or evaluation.
It supports the proposed mechanism-aware completion strategy but does not imply identification of the observation process itself.


\subsection{Semi-Synthetic fMRI Dataset}
The fMRI benchmark introduced by \citet{fmri2011network} contains semi-simulated time series whose nodes represent brain regions.
We evaluate networks with $d\in\{5,10\}$ under sample-wise missingness.
Because the benchmark provides only lagged ground-truth edges, this experiment focuses exclusively on recovery of $\mathbf{B}$.
Therefore, we include the Granger causality-based method \textbf{CUTS+}~\citep{cuts+2024cheng} in this experiment.
However, this benchmark does not verify every ITSY assumption. 
Thus, the results provide graph-recovery evidence rather than neuroscientific validation.

\begin{table}[t]
\centering
\caption{Results on the semi-synthetic fMRI benchmark with sample-wise missingness.}
\label{tab2}


\begin{tabular}{lcccc@{\hspace{8pt}}lcccc}
\toprule
\multicolumn{5}{c}{$p=0.3$}
&
\multicolumn{5}{c}{$p=0.5$}
\\
\cmidrule(lr){1-5}
\cmidrule(lr){6-10}

& \multicolumn{2}{c}{5 Regions}
& \multicolumn{2}{c}{10 Regions}
&
& \multicolumn{2}{c}{5 Regions}
& \multicolumn{2}{c}{10 Regions}
\\
\cmidrule(lr){2-3}
\cmidrule(lr){4-5}
\cmidrule(lr){7-8}
\cmidrule(lr){9-10}

Method & F1 & SHD & F1 & SHD
& Method & F1 & SHD & F1 & SHD
\\
\midrule

ITSY
& \textbf{0.62} & \textit{7.06}
& \textbf{0.58} & \textit{14.10}
& ITSY
& \textbf{0.61} & \textbf{7.94}
& \textbf{0.48} & \textit{19.6}
\\

DYNOTEARS
& 0.45 & 8.12
& 0.32 & 17.20
& DYNOTEARS
& 0.12 & 9.35
& 0.05 & 20.4
\\

NTS-NOTEARS
& 0.58 & 11.59
& 0.38 & 64.80
& NTS-NOTEARS
& 0.56 & 10.59
& 0.40 & 50.6
\\

PCMCI+
& 0.43 & \textbf{6.88}
& 0.53 & \textbf{13.40}
& PCMCI+
& 0.28 & 8.06
& 0.31 & \textbf{16.8}
\\
Rhino
& 0.44 & 13.56
& 0.35 & 28.57
&Rhino
& 0.34 & 12.6
& 0.29 & 22.7
\\
CUT+ (Granger)
& 0.28 & 10.5
& 0.30 & 18.67
& CUT+ (Granger)
& 0.22 & 20.3
& 0.25 & 23.5
\\
\bottomrule
\end{tabular}
\end{table}
ITSY achieves the highest F1 in all settings in Tab.\ref{tab2}, while its SHD remains close to PCMCI+ and below the other baselines.
PCMCI+ obtains slightly lower SHD in all settings, showing that a sparse estimate may require fewer edits while recovering fewer true edges.
Meanwhile, CUTS+ shows no clear advantage when the data contain missing values.
The combination of higher F1 and competitive SHD indicates improved recovery without a disproportionate increase in additional edges.
These results transfer the synthetic findings to a distinct simulator, although the network sizes and simulator-specific hemodynamics preclude neuroscientific claims.

\subsection{Scope of evidence}
Our evidence is limited to stationary, causally sufficient, linear first-order SVARs under MCAR.
The analysis supports the formulation, while synthetic experiments show improved graph recovery across missing rates, dimensions, and graph densities.
The semi-synthetic fMRI study tests transfer to a different simulator, but only on small networks with lagged ground-truth graphs.
We do not evaluate informative missingness, asynchronous measurement clocks, latent confounding, or time-varying mechanisms.
These results therefore do not establish performance on naturally irregular clinical data or support neuroscientific conclusions.

\section{Conclusion}
\label{conclusion}
We introduce ITSY for jointly imputing irregular observations and recovering contemporaneous and lagged graphs in linear first-order SVARs.
Although our setting is simple, it is rigorously formulated and supported by comprehensive theoretical analysis.
By predicting from the nearest observed history and optimizing a weighted structural reconstruction objective, ITSY improves recovery across synthetic and a semi-synthetic benchmark.
The current evidence is bounded by assumptions. 
In particular, the missing mask is sampled independently of observed and unobserved values.
Future work will examine MAR and MNAR processes, in which missingness may depend on observed or latent variables, and will extend the structural estimator to nonlinear mechanisms and higher-order SVARs with latent confounders.
These extensions require new identification and optimization arguments. 

\clearpage


\bibliographystyle{plainnat}
\bibliography{ref}
\clearpage

\maketitle
\appendix
{\bf\Large\centering Appendix \\[1em]}
The appendix is organized as following sections:
\begin{itemize}[leftmargin=9pt]
\item \textbf{A. Theory assumptions and proofs} provides necessary proofs for Lemma 1, which concerns the formal expression of the differential constraints of the DAG; and Lemma 2, Gauss-Markov Theorem.
\item \textbf{B. Model Details} provides details of proposed estimator, convergence analysis for ITSY and time consumptions across different scales and missing rate. 
\item \textbf{C. Experiment Details} describes the details of baseline models as well as experimental parameters, including model hyper-parameters, generating process and metrics.
\item \textbf{D. Additional Results} provides additional experiments results, in the line with main paper but different variable counts. These illustrates the robustness of ITSY.
\end{itemize}
\section{assumptions and proofs}
\label{A_A_proof}
\subsection{Prove $(\mathbf{I}-\mathbf{A})$ is invertible}
Let $\mathbf{A}\in\mathbb{R}^{d\times d}$ be the weighted adjacency matrix of a DAG.
Then $\mathbf{I}-\mathbf{A}$ is invertible, and:
\begin{equation}
(\mathbf{I}-\mathbf{A})^{-1}
=
\sum_{k=0}^{d-1}\mathbf{A}^{k}.
\end{equation}

\textit{Proof.}
Because $\mathbf{A}$ represents a DAG, there exists a permutation matrix $\mathbf{P}$ corresponding to a topological ordering such that
$\mathbf{PAP}^{\top}$ is strictly triangular.
Therefore, $\mathbf{A}$ is nilpotent and satisfies $\mathbf{A}^{d}=\mathbf{0}$.
Consequently,
\begin{equation}
(\mathbf{I}-\mathbf{A})
\left(
\mathbf{I}+\mathbf{A}+\cdots+\mathbf{A}^{d-1}
\right)
=
\mathbf{I}-\mathbf{A}^{d}
=
\mathbf{I}.
\end{equation}
Thus, $\mathbf{I}-\mathbf{A}$ is invertible, with inverse
$\sum_{k=0}^{d-1}\mathbf{A}^{k}$.
\hfill$\square$

\subsection{Proof of Lemma 1}
\label{proof1}
According to \citet{NOTEARS2018dags}, suppose $B \in \{0,1\}^{d \times d}$, $\operatorname{tr} B^k$ counts the number of length-$k$ closed walks in a directed graph. 
Then $B$ has no cycles if and only if:
 \begin{equation}
    \sum_{k=1}^{\infty} \sum_{i=1}^{d} (B^k)_{ii} = 0
 \end{equation}
From power series of matrix exponential, one can derive:
\begin{align}
    \operatorname{tr} e^B &=\operatorname{tr}\sum_{k=0}^{\infty} \frac{1}{k!}B^k \nonumber \\
    &=\operatorname{tr}I + \operatorname{tr}\sum_{k=1}^{\infty}\frac{1}{k!}B^k \nonumber \\
    &=d + \sum_{k=1}^{\infty}\sum_{i=1}^{d}  \frac{1}{k!}(B^k)_{ii}/k! \nonumber \\
    &=d
\end{align}

To extend this to matrices with an arbitrary weighted matrix $W$ with both positive and negative values, one can simply use the Hadamard product $W \circ W$, which leads to:

A matrix $W \in \mathbb{R}^{d \times d}$ is a DAG if and only if
\begin{equation}
h(W) = \operatorname{tr}\left(e^{W \circ W}\right) - d = 0,
\end{equation}
where $\circ$ is the Hadamard product and $e^A$ is the matrix exponential of $A$.

\subsection{Proof of Lemma 2}
\label{proof2}
Let \(\tilde{\boldsymbol\beta} = \mathbf A\,\mathbf y\) be any linear unbiased estimator, in which:
\begin{align}
   & \mathbb{E}[\tilde{\boldsymbol\beta}]
= \mathbf A\,\mathbb{E}[\mathbf y]
= \mathbf A\,(\mathbf X\boldsymbol\beta)
= \boldsymbol\beta, \nonumber\\
\Rightarrow &\mathbf{AX} = \mathbf{I}
\end{align}

Define $\mathbf{C}=\mathbf{A}-(\mathbf{X^\top\mathbf X)^{-1} X^\top}$,
\begin{equation}
    \mathbf{CX} =\mathbf{AX}- (\mathbf{X^\top\mathbf X)^{-1}X^\top X}
    = \mathbf{AX}-\mathbf{I}
    = 0
\end{equation}
The estimator of OLS defined as $\hat{\boldsymbol\beta}=(\mathbf{X^\top\mathbf X)^{-1} X^\top y}$, then:
\begin{align}
    \tilde{\boldsymbol\beta} &= \mathbf A\,\mathbf y \nonumber\\
&=\mathbf{[(X^\top\mathbf X)^{-1} X^\top +C]}\ \mathbf{y} \nonumber\\
&= \hat{\boldsymbol\beta} + \mathbf{Cy} \nonumber\\
&= \hat{\boldsymbol\beta} + \mathbf{C(\mathbf X\boldsymbol\beta+\epsilon)} \nonumber\\
&= \hat{\boldsymbol\beta} + \mathbf C\boldsymbol\varepsilon 
\end{align}

\(\hat{\boldsymbol\beta}\) and \(\mathbf C\,\boldsymbol\varepsilon\) are both linear in \(\boldsymbol\varepsilon\) and \(\mathbf C\mathbf X=0\), hence they are uncorrelated,  $\text{Cov}(\hat{\boldsymbol\beta},\mathbf{ C\,\boldsymbol\varepsilon})=0$.

Then:
\begin{align}
\mathrm{Var}(\tilde{\boldsymbol\beta}) &= \mathrm{Var}(\hat{\boldsymbol\beta} + \mathbf C\boldsymbol\varepsilon )\nonumber \\
&= \mathrm{Var}(\hat{\boldsymbol\beta}) + \mathrm{Var}(\mathbf C\,\boldsymbol\varepsilon) + 2\text{Cov}(\hat{\boldsymbol\beta},\mathbf{ C\,\boldsymbol\varepsilon})\nonumber \\
&= \mathrm{Var}(\hat{\boldsymbol\beta}) + \sigma^2\,\mathbf C\,\mathbf C^\top
\end{align}

Because \(\mathbf C\,\mathbf C^\top\) is positive semidefinite, $ \mathrm{Var}(\tilde{\boldsymbol\beta})\succeq \mathrm{Var}(\hat{\boldsymbol\beta})$

This shows \(\hat{\boldsymbol\beta}\) has the smallest variance among all linear unbiased estimators.

\section{Model Details}
\label{A_B_coverge}
\paragraph{Detailed Time and Space Complexity}
Let $d$ denote the number of variables, $T$ the sequence length, and $n$ the number of samples or trajectories.
The transition operator $\mathbf{P}_{\theta}$ is obtained by solving the dense linear system $(\mathbf{I}-\mathbf{A})\mathbf{P}_{\theta}=\mathbf{B}$, which requires $O(d^3)$ time with a standard dense solver.
Once $\mathbf{P}_{\theta}$ is available, each propagation step computes a matrix--vector product $\mathbf{P}_{\theta}\mathbf{X}(t-1)$, incurring $O(d^2)$ time.
Applying this propagation over $T$ time steps and $n$ samples therefore requires $O(Tnd^2)$ time.
Combining the linear solve and temporal propagation gives an overall time complexity of $O(d^3+Tnd^2)$ for one coupled sweep.

During differentiation, the intermediate completed states must be retained in the computation graph, requiring $O(Tnd)$ memory for the propagated states.
Including the model matrices introduces an additional $O(d^2)$ storage term, so the full space complexity is $O(Tnd+d^2)$, which reduces to $O(Tnd)$ when $Tn$ dominates $d$.

\paragraph{Discussion about Estimation}
The objective of OLS is to minimize the sum of squared residuals, which are estimates of the noise $\mathbf{e}(t)$. 
The OLS estimator's efficiency and inference validity are closely tied to the noise properties, see \textit{Lemma 2}.

Let's slightly adjust estimator and compare two equations:
\begin{align}
&{\mathbf{X}}(t) =(\mathbf{I}-\hat{\mathbf{A}})^{-1}  {\mathbf{B}}  \mathbf{X}(t-1) +(\mathbf{I}-{\mathbf{A}})^{-1}\mathbf{e}(t) \nonumber\\
 &\hat{\mathbf{X}}(t) = (\mathbf{I}-\hat{\mathbf{A}})^{-1}  \hat{\mathbf{B}}  \mathbf{X}(t-1)  
\end{align}
The noise item or the estimation residual becomes $(\mathbf{I}-{\mathbf{A}})^{-1}\mathbf{e}(t)$, thus we have the WLS Equation for decoupling, allowing recovery of the residuals as $\mathbf{e}(t)$. 
Note that although this modeling approach eliminates dependence on the $\mathbf{X}(t)$ during estimation,it still requires $\mathbf{X}(t)$ for loss calculation. 
Thus, it essentially models continuous observation samples. 
This explains why our method shows good robustness when there's no missing data.
Assuming there are three steps of time-series $\mathbf{X}(t),\mathbf{X}(t-1),\mathbf{X}(t-2)$, where $\mathbf{X}(t-1)$ are missing.
According to main paper, we have the estimator:
\begin{equation}
    \hat{\mathbf{X}}(t) =[(\mathbf{I}-\hat{\mathbf{A}})^{-1}\hat{\mathbf{B}}]^2  \mathbf{X}(t-2)
\end{equation}
Let $\mathbf{W}=(\mathbf{I}-{\mathbf{A}})^{-1}\mathbf{B}$, the corresponding true generation process is: 
\begin{align}
    {\mathbf{X}}(t) &=\mathbf{W} \mathbf{X}(t-1) + (\mathbf{I}-{\mathbf{A}})^{-1}\mathbf{e}(t)\nonumber \\
    &= \mathbf{W} [ \mathbf{W}  \mathbf{X}(t-2) +(\mathbf{I}{\mathbf{A}})^{-1}\mathbf{e}(t-1)]+ \nonumber \\
    &\quad\quad(\mathbf{I}-{\mathbf{A}})^{-1}\mathbf{e}(t)\nonumber \\   
    &=\mathbf{W}^2\mathbf{X}(t-2) +\mathbf{W}(\mathbf{I}-{\mathbf{A}})^{-1}\mathbf{e}(t-1)+ \nonumber \\
    &\quad\quad(\mathbf{I}-{\mathbf{A}})^{-1}\mathbf{e}(t)
\end{align}
More generally, if the nearest previous complete observation is at time $(t-i)$, then we can obtain the expression:
\begin{equation}
    \mathbf{X}(t)= \mathbf{W}^i {\mathbf{X}}(t-i) + \sum_{j=0}^{i-1} \mathbf{W}^{j} (\mathbf{I}-{\mathbf{A}})^{-1}\mathbf{e}(t-j).
\end{equation}

From the above derivation, it seems that the reweighting \((\mathbf{I} - \mathbf{A})\) term cannot fully capture and recover the noise component.
Nevertheless, this design is still crucial both for observed and irregular time series.
Since missing entries lack real observations, we cannot directly compute residuals to minimize loss.
Therefore, accurate continuous observed samples estimations is crucial, which can help learn the causal mechanisms of generating process.
With the accurate reconstruction of $\mathbf{A}$ and $\mathbf{B}$, imputation of nearby missing samples can benefit from them.
Iteratively, this further promotes estimating subsequent observations, and in return, accurate reconstruction enhances the accuracy of the imputation process.
To this end, we finish the joint optimization of reconstruction and imputation.

On the other hand, we find that full  recovery of $\sum_{j=0}^{i-1} \mathbf{W}^{j} (\mathbf{I}-{\mathbf{A}})^{-1}\mathbf{e}(t-j)$ is computationally demanding in our engineering practice.
As $j$ accumulates, the noise roughly converges to $\textbf{e}(t)$.
Hence, from an efficiency--accuracy trade-off perspective, weighting by $\mathbf{I}-\mathbf{A}$ is sufficient.

\begin{figure}[t] 
    \centering 
    \includegraphics[width=0.5\textwidth]{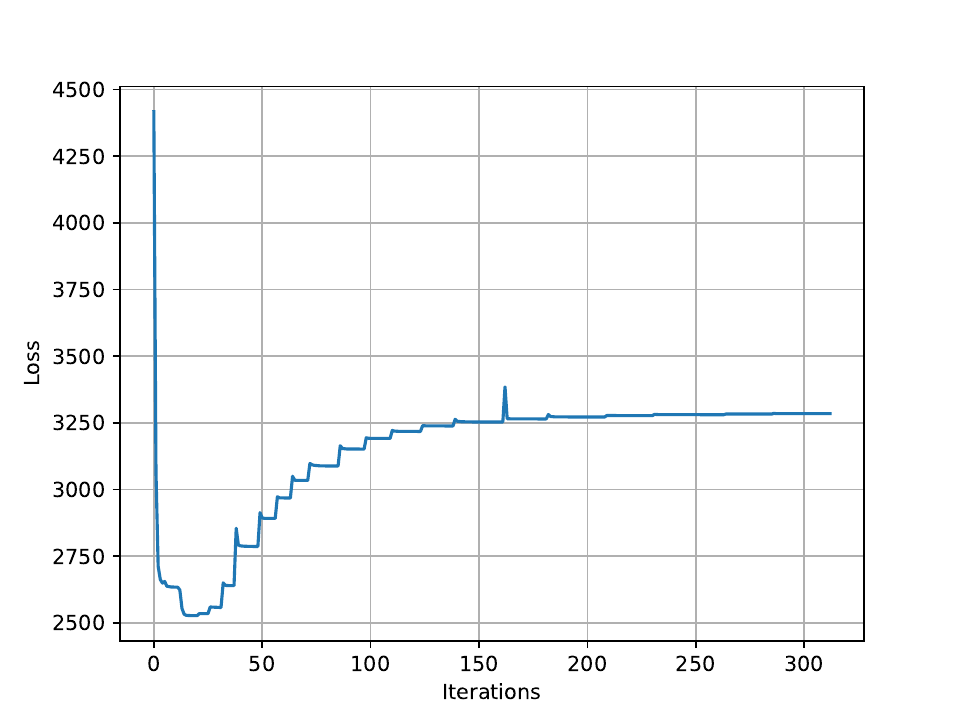}
    \caption{Illustration of model convergence
    } 
    \label{fig4} 
\end{figure}

\paragraph{Convergence Analysis}
We use a $\mathbf{X}\in5000 \times 1 \times 10$ example to record the model's convergence, sampling the loss value every 10 iterations.
As the number of model iterations increases, the loss function value gradually decreases. 
As the iteration nears its end, the loss value stabilizes, indicating model convergence.
See Fig.\ref{fig4} for illustration.




\paragraph{Time Consumption}
\begin{figure}[t] 
    \centering 
    \includegraphics[width=0.45\textwidth]{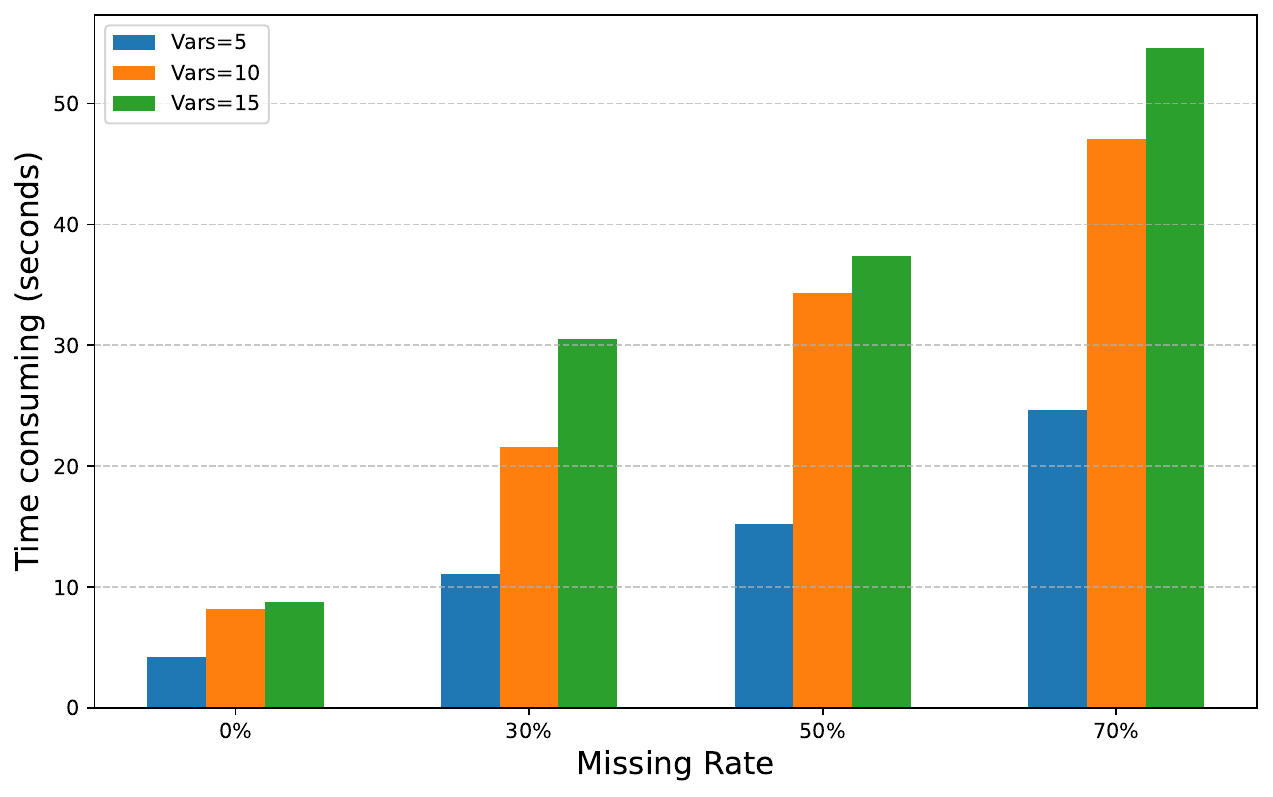}
    \caption{It illustrates the model's time consumption across different variable scales ($d=5,10,15$) as the missing rate increases ($0\%,30\%,50\%,70\%$).Colors of lines represent variable sizes. Each subgroup represents a different missing rate.} 
    \label{figtime} 
\end{figure}
Fig.\ref{figtime} illustrates the model's time consumption across different variable scales as the missing rate increases.
Here, we run the experiments only with observations $T=5000, n=1$.
Nevertheless, other dimension scales maintain the similar trends.
The results show that an increase in missing rates will result in longer convergence times.
As the number of variables increases, there is no exponential increase in time consumption due to combinatorial explosion.
Both time consumptions exhibit a trend of linear growth, reflecting the time robustness of our method.

\section{ Experiment Details}
\label{A_C_baseline}
\subsection{Setup}
We average 50 independent repetitions for each setting. Synthetic data follow a first-order linear SVAR; the semi-synthetic evaluation uses an fMRI benchmark \citep{fmri2011network}. 
Both have shape $\mathbf{X}\in\mathbb{R}^{T\times n\times d}$. 
Experiments use an Intel Xeon Gold 6138 CPU, 32GB memory, and a Tesla V100-PCIE-32GB GPU.

\subsection{Experiment Details}
All the baseline works use the default hyper-parameters.
We only make necessary modifications to accommodate the input data.
Following the common backbone, MLP-NOTEARS \citep{zheng2020NOTEARSMLP}, all the continuous optimization methods using Wrap L-BFGS-B optimizer, using scipy routines. \footnote{https://gist.github.com/arthurmensch}

\subsubsection{ITSY}
\paragraph{Engineering}
The implementation of this work is based on YLearn \citep{YLearn} and NOTEARS-MLP \citep{zheng2020NOTEARSMLP}.
We rewrite the forward function and model structure, adjust the objective function, and switch to a more efficient optimizer in optimization process, while keep the original generation mechanism.
\paragraph{Hyper-parameters} For sparsity constraint,
\begin{align*}
    \lambda_A=0.01,
    \lambda_B=0.01
\end{align*}
For initializing Augmented Lagrangian multiplier in DAG constraint,
\begin{align*}
    &\rho=1.0, \rho_{max}=1e16,\\
    &\alpha=0.
\end{align*}
For training and optimization,
\begin{align*}
    & optimizer: \text{Wrap L-BFGS-B algorithm, using scipy routines}\\
    & threshold=0.3,\\
    & max\_epochs=20,\\
    & learning\_rate=0.01,\\
    & DAG\_bounds: [0,0] \ \text{for diagonals and [-1,1] for others} 
\end{align*}

\subsubsection{DYNOTEARS}
\paragraph{Details} 
DYNOTEARS is proposed by \citet{pamfil2020dynotears}, it's the first SCM-based work for temporal causal discovery with continuous optimization.
DYNOTEARS bases linear SVAR and offers theoretical analysis and empirical evidence.
Thus, we select it as one of the comparison baselines.

Due to the absence of official code, we opted for YLearn's implementation \citep{YLearn} and enhanced it by refining the lagged item computation and optimizer, drawing on the NOTEARS-MLP approach \citep{zheng2020NOTEARSMLP}. 
This allowed us to boost computational efficiency without compromising results.
As illustrated in experiments, it delivers the superior performance under full observations.

\paragraph{Hyper-parameters} For sparsity constraint,
\begin{align*}
    \lambda_A=0.01,\\
    \lambda_B=0.01
\end{align*}
For initializing Augmented Lagrangian multiplier in DAG constraint,
\begin{align*}
    &\rho=1.0, \rho_{max}=1e16,\\
    &\alpha=0.
\end{align*}
For training and optimization,
\begin{align*}
& optimizer: \text{Wrap L-BFGS-B algorithm, using scipy routines}\\
    & threshold=0.3,\\
    & max\_epochs=20,\\
    & learning\_rate=0.01,\\
    & DAG\_bounds: [0,0] \ \text{for diagonals and [-1,1] for others} 
\end{align*}
\subsubsection{NTS-NOTEARS} 
\paragraph{Details} 
NTS-NOTEARS is proposed by \citet{sun2023nts}.
It's also a continuous optimization work.
It utilizes 1D convolutional neural networks (1D-CNNs) to model the dependence of child variables on their parents.
They demonstrate the following theorem: Let \( F \) be the class of 1D-CNNs that are independent of \( X_{i}^{k} \) and \( F_{0} \) be the class of 1D-CNNs such that the \( i \)-th kernel parameters in the \( k \)-th column of the \( m \) first-layer CNN kernels are all zeros. Then, \( F = F_{0} \).

NTS-NOTEARS is designed for nonlinear settings, but the authors state that it performs well in linear settings.
We modified several default settings in the code to adapt it to the linear mechanism of \(\mathbf{X}\in \mathbb{R}^{T \times n\times d} \) and high-dimensional inputs.
We first remove data normalization as it can disrupt the noise variance, as stated in \textit{Lemma 2}.
We then leverage the batching capability of neural networks by treating \(\mathbf{X} \in \mathbb{R}^{T \times n \times d}\) as \(n\) batches of \(\mathbf{X} \in \mathbb{R}^{T \times d}\) to ensure a fair experimental comparison.
\paragraph{Hyper-parameters} For sparsity constraint, they utilize both $l_1-Norm$ and $l_2-Norm$,
\begin{align*}
    \lambda_1&=0.005,\\
    \lambda_2&=0.01
\end{align*}
For initializing Augmented Lagrangian multiplier in DAG constraint,
\begin{align*}
    &\rho=1.0, \rho_{max}=1e16,\\
    &\alpha=0.
\end{align*}
For 1D-convolution kernel,
\begin{align*}
   &  int\_channel=d,\\
   &  out\_channels=d \times 10,\\
   &  bias=True,\\
   &  kernel\_size=lags+1,\\
   &  stride=1, \\
   &  padding=0
\end{align*}
For training and optimization,
\begin{align*}
& optimizer: \text{Wrap L-BFGS-B algorithm, using scipy routines}\\
    & threshold=0.3,\\
    & learning\_rate=0.01,\\
    & max\_iterations=100,\\
    & DAG\_bounds: [0,0] \ \text{for diagonals and [0,None] for others} 
\end{align*}

\subsubsection{PCMCI+} 
\paragraph{Details}
PCMCI+ is proposed by \citet{pcmci+2020runge}, provides a two-stage, conditional-independence framework for reconstructing causal graphs from multivariate time series. 
Its core innovation is the Momentary Conditional Independence (MCI) test, which adjusts for autocorrelation by conditioning on both the candidate cause's and effect's estimated parents.
This tailored conditioning yields well-calibrated $p$-values at nominal false-positive rates---even in strongly autocorrelated data---and boosts detection power relative to standard conditional-independence tests.

\paragraph{Hyper-parameters}
For high-dimensional data pre-processing, we choose $$analysis\_mode=\text{"multiple"}$$
$$significance=\text{"analytic"}$$
\begin{align*}
    &tau\_min = 1,tau\_max = 3\\
    &pc\_alpha = 0.01
\end{align*}

\subsubsection{Rhino}
\paragraph{Details}
Rhino is proposed by \citet{gong2024rhino} as a deep generative framework for temporal causal discovery.
It models nonlinear temporal relationships with both instantaneous and lagged causal effects, while allowing the exogenous noise distribution to depend on historical observations.
Rhino jointly learns a distribution over temporal causal graphs and the associated functional relationships, and provides structural identifiability under its stated assumptions.

\paragraph{Hyper-parameters}
Following the official implementation, we configure Rhino with the maximum temporal lag used in our benchmark and employ Gaussian exogenous noise.
The main settings are
\begin{align*}
    &base\_distribution\_type = \text{"gaussian"},\\
    &allow\_instantaneous = \text{True},\\
    &lag = 1.
\end{align*}

\subsubsection{CUTS+}
\paragraph{Details}
CUTS+ is proposed by \citet{cuts+2024cheng} for causal discovery from high-dimensional irregular time series.
Building upon the Granger-causality-based CUTS framework, it alternates between causal graph learning and missing-data imputation, while introducing Coarse-to-Fine Discovery (C2FD) to reduce the graph search space.
It further employs a message-passing graph neural network (MPGNN) for data prediction, reducing structural redundancy and improving scalability to high-dimensional time series with irregular observations.

\paragraph{Hyper-parameters}
Following the official implementation, we use the reported CUTS+ configuration:
\begin{align*}
    &batch\_size = 128,\quad hidden\_size = 32,\\
    &Gumbel\ \tau: 1 \rightarrow 0.1,\\
    &\lambda = 0.01.
\end{align*}
The learning rates are annealed during the two alternating stages following the implementation.

\clearpage
\subsection{Generating Process}
This research follows the same generating process alongside with YLearn's implementation \citep{YLearn}.
The generating process takes two steps: graph generating and time-series generating.
\subsubsection{Graph Generating}
We use directed Erd\H{o}s--R\'enyi graphs for graph generation.
In the directed Erd\H{o}s--R\'enyi graph $\vec G(d,p)$, there are $d$ vertices and each of the $d(d-1)$ ordered pairs of distinct vertices is included independently with probability~$p$, so the expected number of directed edges is:
\begin{equation}
    \mathbb{E}\bigl[|E_{\to}|\bigr] \;=\; p\,d(d-1).    
\end{equation}

Each vertex $v$ has indegree and outdegree distributed as $\mathrm{Bin}(d-1,p)$ and total degree $\deg(v)=\deg^{\mathrm{in}}(v)+\deg^{\mathrm{out}}(v)$ has expectation $2p(d-1)$. 
Moreover, since both the average degrees satisfy $\tfrac{|E_{\to}|}{d}=p(d-1)$, and the average total degree is $\tfrac{2|E_{\to}|}{d}=2p(d-1).$

In the implementation, we first specify the degree to generate an adjacency matrix, then create a weighted adjacency matrix within a given value range: $(-0.95, -0.5)\,\cup\,(0.5, 0.95)$.
\subsubsection{Time-series Generating}
After generating the weighted graph, we can subsequently generate time series.
We first initialize a \( \mathbf{X}\in \mathbb{R}^ {k\times n \times d} \) random variable as the initial observations, and then iteratively generate $T$ samples based on the predefined structural equations Eq.1.
Thus we have $\mathbf{X}\in \mathbb{R}^{(T+k)\times n\times d}$.
Finally, we drop the samples \( \mathbf{X}[0; \text{k}] \), as initial observations cannot be used for estimation.

\subsection{Metrics}

The code for evaluation metrics is either adopted from or adapted based on NOTEARS \citep{NOTEARS2018dags}. 
For contemporaneous matrices, we utilize the metrics provided by NOTEARS. 
For lagged matrices, we have adjusted and optimized the input metrics provided by NOTEARS.
Since the metrics are designed for DAGs, the rows and columns must represent the same variable.
However, for lagged graphs, rows and columns represent variables at different time points \( t-1 \) and \( t \), respectively. 
The computational logic provided in the original work can lead to misspecification.

\begin{figure}[h] 
    \centering 
    \includegraphics[width=0.8\textwidth]{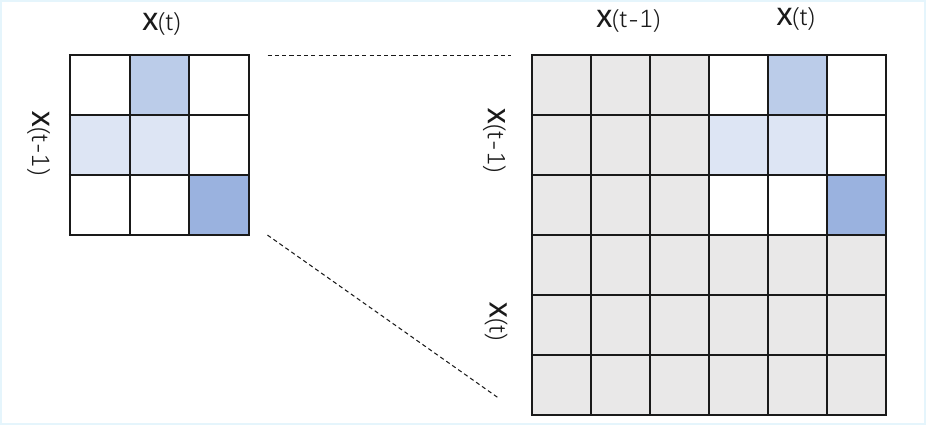}
    \caption{Illustration of matrix expansion
    } 
    \label{fig5} 
\end{figure}

To address this, we incorporate the true and estimated values of \( \textbf{B}_1 \) into a larger matrix of size \( 2d \times 2d \), where the horizontal and vertical coordinates represent variables at time \( t-1 \) and \( t \) respectively, within the ranges \([0, d)\) and \([d, 2d)\). 
Consequently, \( \textbf{B}_1 \) corresponds to the upper right submatrix of this matrix.
Although the scale is larger, the metric calculations will focus only on the relevant portions that contain the \( \textbf{B}_1 \) information. 
This approach ensures that the evaluation metrics are correctly applied to the lagged graph structure without misspecification.
See Fig.\ref{fig5} for illustration.

\clearpage
\section{Additional Results}
\label{A_D_Exp}
\begin{figure*}[h] 
    \centering 
    \includegraphics[width=\textwidth]{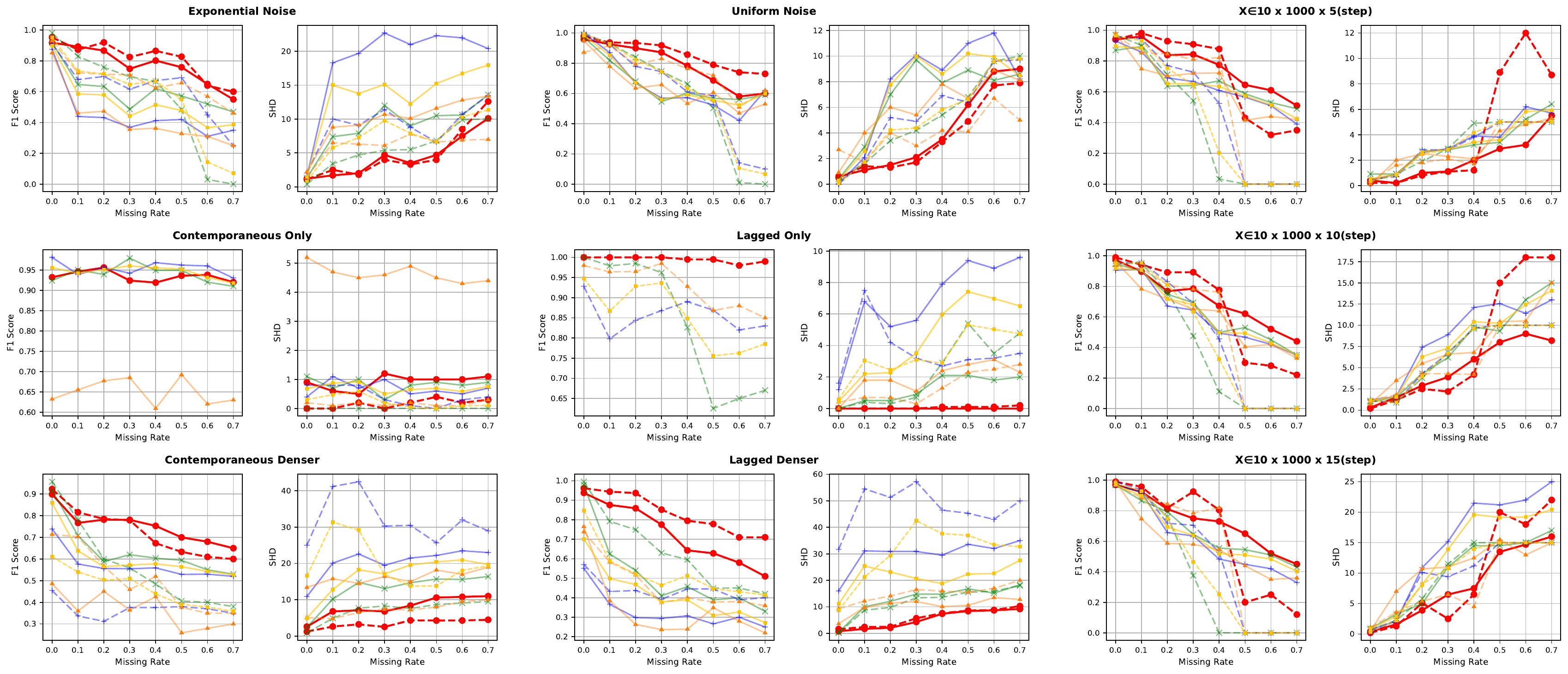}
    \includegraphics[width=0.7\textwidth]{material/bar.pdf}
    \caption{Performance of different generation settings. 
    The noise distribution and sparsity are changed.
    ``Only'' means alternated graph is empty and ``Denser'' means more edges.
    Note that the F1-scores for the empty graphs are undefined and SHDs only indicate extra edges.
    As for the 3rd column, we apply Step-wise Missing for high dimension setting
} 
    \label{fig3} 
\end{figure*}
\begin{figure*}[htbp] 
    \centering 
    \includegraphics[width=\textwidth]{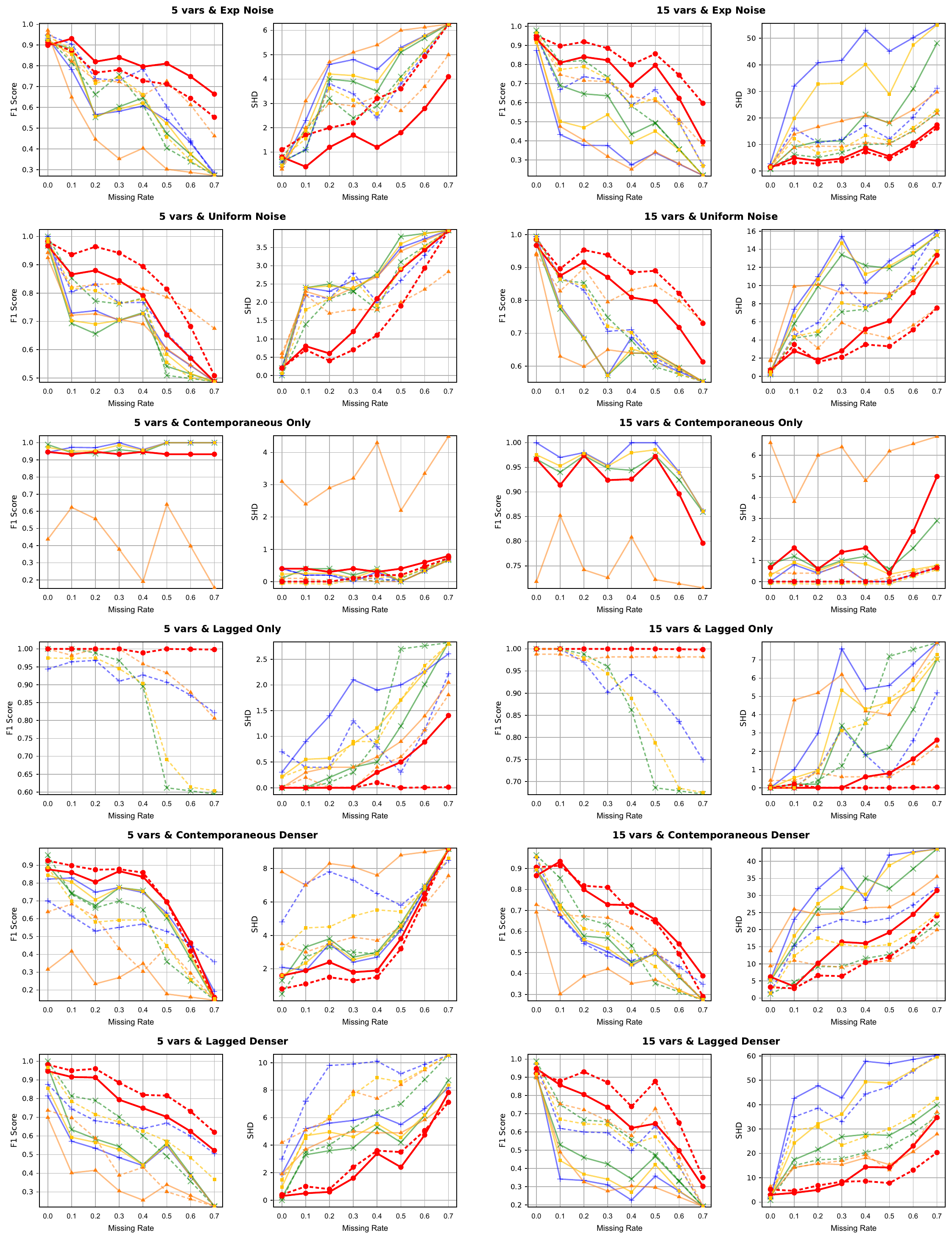}
        \includegraphics[width=0.7\textwidth]{material/bar.pdf}
    \caption{Additional results across varying generation settings, supplement for Fig.\ref{fig3}.
    Following the different noise distribution and density, we try more and less variables counts of 5 and 15.
    } 
    \label{fig7} 
\end{figure*}

Fig.\ref{fig3}, Fig.\ref{fig7} illustrate additional results across varying generation settings.
In high dimension setting with Step-wise Missing (Fig.\ref{fig3}, the 3rd column), all four models  maintain the same degradation trend as in Tab.1 (See main paper): achieving brilliant results in full observations and get totally failed when $70\%$ missing rate, since the observations degrades into subsampling or even worse.
However, between $10\%$ and $50\%$, ITSY shows its effectiveness and superiority.
The less variables means the less complex dependency and vice versa for more variables.
In both settings, ITSY consistently outperforms all baselines across nearly every missingness level and remains stable up to 50\% missing data, demonstrating its effectiveness and correctness in irregular settings.

Overall, Under full observations, all the methods shows excellent performance, and ITSY shows its effectiveness and outperform other baseline with missing samples.
NTS-NOTEARS exhibits sensitivity to sample size and variable counts, while PCMCI+ in a few settings nearly matches ITSY.


\end{document}